\documentclass[11pt,a4paper]{article}
\usepackage{acl2021}
\usepackage{times}
\usepackage{latexsym}

\usepackage{amsmath,amsfonts,bm}

\def\eqref#1{equation~\ref{#1}}

\def\1{\bm{1}}

\DeclareMathAlphabet{\mathsfit}{\encodingdefault}{\sfdefault}{m}{sl}
\SetMathAlphabet{\mathsfit}{bold}{\encodingdefault}{\sfdefault}{bx}{n}

\usepackage[utf8]{inputenc} 
\usepackage[T1]{fontenc}    
\usepackage{hyperref}       
\usepackage{url}            
\usepackage{booktabs}       
\usepackage{amsfonts}       
\usepackage{nicefrac}       
\usepackage{microtype}      
\usepackage{amsmath,amssymb}
\usepackage{epsfig,graphicx,subfigure,caption}
\usepackage{algpseudocode}
\usepackage{multirow,booktabs, hhline}
\usepackage[ruled,noend]{algorithm2e}
\usepackage{amsmath, bm}
\usepackage{wrapfig}
\usepackage{lipsum}
\usepackage{caption}
\usepackage{footnote}
\usepackage[bottom]{footmisc}
\usepackage[toc,page]{appendix}

\newcommand{\ronan}[1]{\textcolor{teal}{RL: #1}} 

\usepackage{microtype}
\newcommand{\method}{commonsense knowledge transfer\xspace}
\newcommand{\baby}{CKT\xspace}
\newcommand{\paratitle}[1]{\noindent\textbf{#1}\ \ }

\title{Commonsense Knowledge Transfer for Pre-trained Language Models}

\author{Wangchunshu Zhou\thanks{~~Work done while interning at the Allen Institute for AI}\ \ \textsuperscript{1} \hspace{.3cm} 
\textbf{\hspace{.3cm} Ronan Le Bras} \textsuperscript{2} \hspace{.3cm}
\textbf{\hspace{.3cm} Yejin Choi} \textsuperscript{2~3} \\
\textsuperscript{1}ETH Zurich \\
\textsuperscript{2}Allen Institute for AI \hspace{.3cm}  \\
\textsuperscript{3}Paul G. Allen School of Computer Science \& Engineering, University of Washington \\
\texttt{wangchunshu.zhou@inf.ethz.ch}}

\begin{document}
\maketitle

\begin{abstract}
Despite serving as the foundation models for a wide range of NLP benchmarks, pre-trained language models have shown limited capabilities of acquiring implicit commonsense knowledge from self-supervision alone, compared to learning linguistic and factual knowledge that appear more explicitly in the surface patterns in text. 

In this work, we introduce \textit{commonsense knowledge transfer}, a framework to transfer the commonsense knowledge stored in a neural commonsense knowledge model to a general-purpose pre-trained language model. It first exploits general texts to form queries for extracting commonsense knowledge from the neural commonsense knowledge model and then refines the language model with two self-supervised objectives: \textit{commonsense mask infilling} and \textit{commonsense relation prediction}, which align human language with the underlying commonsense knowledge.

Empirical results show that our approach consistently improves the model's performance on downstream tasks that require commonsense reasoning. Moreover, we find that the improvement is more significant in the few-shot setting. This suggests that our approach helps language models better transfer to downstream tasks without extensive supervision by injecting commonsense knowledge into their parameters.
\end{abstract}
\section{Introduction}

Recent advances in pre-trained language models have transformed the landscape of natural language processing. Self-supervised pre-training objectives including masked language modeling~\citep{DBLP:conf/naacl/DevlinCLT19} and masked span infilling~\citep{DBLP:conf/acl/LewisLGGMLSZ20} enable pre-trained models to acquire linguistic~\citep{DBLP:conf/naacl/HewittM19,DBLP:journals/pnas/ManningCHKL20} and factual knowledge~\citep{DBLP:conf/emnlp/PetroniRRLBWM19} by modeling the distribution of naturally occurring texts. However, most of these objectives are limited to exploiting the surface form of human language, and the lack of grounded supervision calls into question how well these representations can ever capture meaning~\citep{DBLP:conf/acl/BenderK20}, not to mention the underlying commonsense knowledge which is often reasoned implicitly and does not appear in the surface form of human language~\citep{DBLP:journals/corr/abs-2104-10809,DBLP:conf/aaai/ZhouZCH20,DBLP:conf/aaai/HwangBBDSBC21}.
On the other hand, commonsense reasoning is important for building generalizable models because it enables the model to reason about a great number of events, causes, and effects, while observing only a small fraction of them. The ineffectiveness of self-supervised language model pre-training on acquiring commonsense knowledge makes them require a relatively large number of labeled examples to succeed in a downstream task and prune to overfit task-specific correlations~\citep{DBLP:journals/tacl/TuLGH20}. 

\begin{figure}[t]
\centering
\small
     \includegraphics[width=\linewidth]{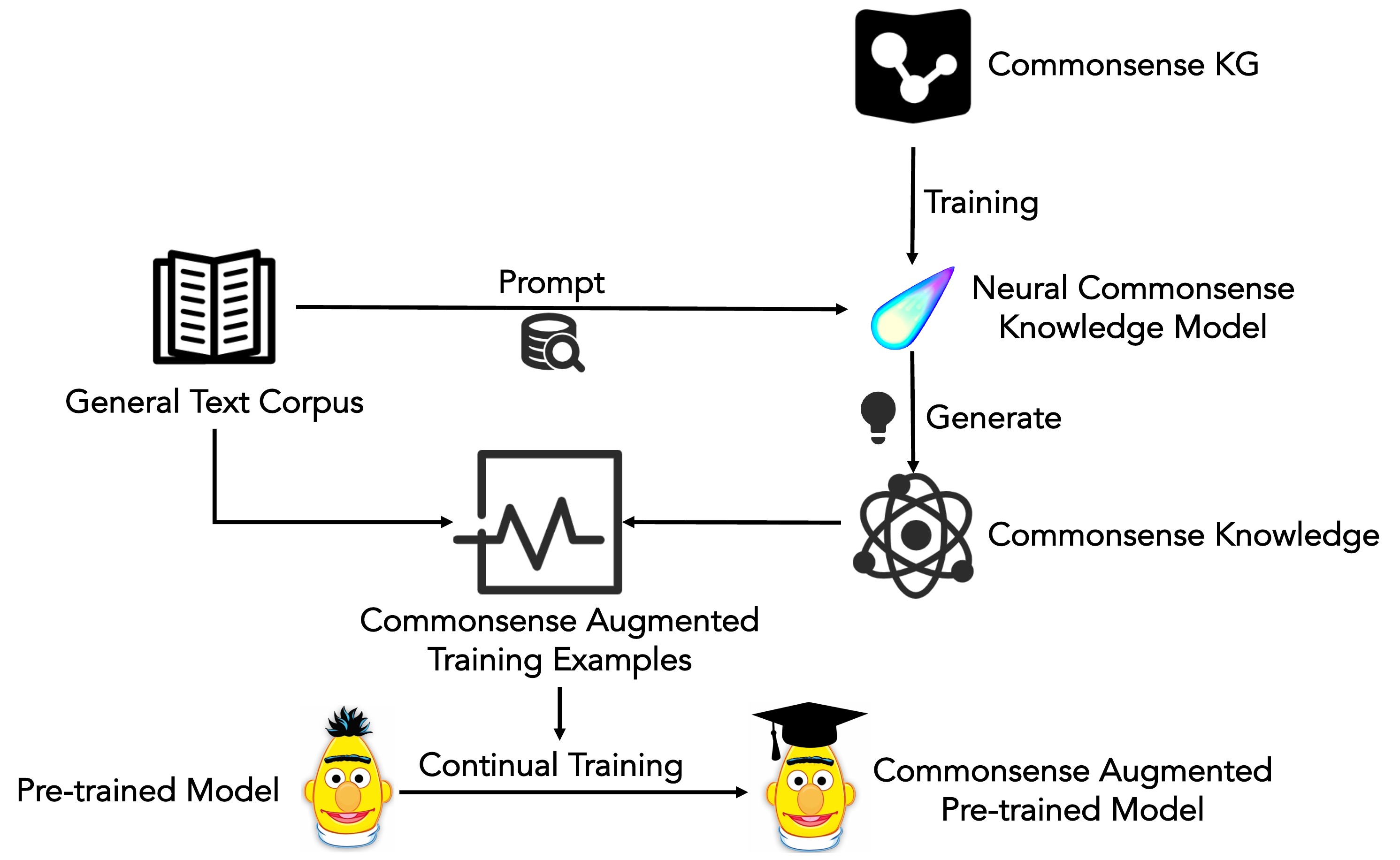} 
     \caption{Illustration of the \method framework. We first extract commonsense knowledge related to sentences in general text corpus from a neural commonsense knowledge model. We then use natural texts and the extracted commonsense knowledge to form self-supervised training data to refine a pre-trained model with commonsense knowledge.}
 \label{fig:framework}
 \vspace{-0.5cm}
 \end{figure}

Therefore, equipping pre-trained language models with commonsense reasoning ability has attracted much attention. To this end, two distinct lines of research focus on improving commonsense reasoning ability of pre-trained language models. The first one focuses on incorporating external commonsense knowledge graph for commonsense reasoning~\citep{DBLP:conf/emnlp/LinCCR19,DBLP:conf/aaai/LiuW0PY21, DBLP:journals/corr/abs-2109-02572} while the other attempts to inject commonsense knowledge into the parameters of pre-trained models~\citep{DBLP:journals/corr/abs-1909-09743,DBLP:conf/iclr/ZhouLSL021,DBLP:journals/corr/abs-2109-05105}. In this work we focus on the second type of method because it alleviates the need for external knowledge bases for training and inference on downstream tasks, thus simpler, more efficient, and not limited by the coverage issue of external knowledge bases.

Prior work injects commonsense knowledge into pre-trained models either on symbolic commonsense knowledge graphs with manually defined rules~\citep{DBLP:journals/corr/abs-1909-09743} or masked language modeling~\citep{DBLP:journals/corr/abs-2112-08615} or on general text corpus with concept-centric self-supervised objectives~\citep{DBLP:conf/iclr/ZhouLSL021}. The former method is limited by the coverage of knowledge graphs and human-written rules. It also fails to make use of large-scale diverse natural text corpus. Therefore, the training is limited to short and synthetic commonsense tuples, which affects its generalization ability on diverse downstream tasks. The latter method, however, only captures surface-level order relations between concepts and fails to learn commonsense relations between concepts such as cause, effect, intent, requirement, etc., which are crucial for commonsense reasoning but often implicitly reasoned, thus do not appear in the surface form of natural language.



In this work, we propose \textit{commonsense knowledge transfer}, an alternative framework to refine a general purpose pre-trained model's commonsense reasoning ability. In contrast to previous work, it aims to transfer the commonsense knowledge stored in a neural commonsense knowledge model (e.g., COMET~\citep{DBLP:conf/acl/BosselutRSMCC19}) to a general purpose pre-trained model on large scale general text corpus. In this way, our approach combines the best of both worlds from prior art: the dense and informative commonsense knowledge from commonsense knowledge graphs and the accessibility of large-scale diverse general corpus.


Commonsense knowledge transfer is conceptually related to knowledge distillation (KD)~\citep{DBLP:journals/corr/HintonVD15} since they both aim to transfer knowledge from a knowledge-rich model to another model that lacks it. However, different from conventional KD, in \method, the source model (i.e., neural commonsense model) and the target model (i.e., pre-trained model) are heterogeneous. Moreover, instead of simply mimicking the teacher model, \method requires the target model to learn specialized knowledge from the source model while retaining its own capability.
This poses unique challenges since the knowledge transfer can not be accomplished by simply matching the logits or feature distribution between the student and the teacher. To this end, we propose to first extract commonsense knowledge in textual form from the source model and then exploit the extracted knowledge to form self-supervised training data for the target model. As illustrated in Figure \ref{fig:framework}, \method first exploits general texts to form queries for retrieving commonsense knowledge from the neural commonsense knowledge model. Then it refines a pre-trained model with two self-supervised objectives that align the surface form of human language with its underlying commonsense inference: \textit{commonsense text infilling} and \textit{commonsense relation prediction}. The former objective concatenates natural text with its commonsense inference to form an input example, masks certain spans in it, and trains the model to reconstruct the original input. The latter method instead trains the model to distinguish valid commonsense inference from carefully constructed spurious commonsense inference given the original text and commonsense relation. Refining a pre-trained model by multi-tasking on both generation (former) and understanding (latter) tasks enables the model to better adapt to different kinds of downstream tasks.

We refine T5~\citep{DBLP:journals/jmlr/RaffelSRLNMZLL20} with \method and fine-tune the resulting model downstream tasks requiring commonsense reasoning ability in both the fully supervised setting and few-shot settings where only a percentage of labeled examples are available. Experimental results show substantial improvements in downstream tasks requiring commonsense reasoning, especially in the few-shot setting, demonstrating the effectiveness of our approach.
\section{Methodology}

Our proposed \method framework consists of a neural commonsense knowledge model (e.g., COMET) and a pre-trained model (e.g., T5). The goal of \method is to transfer the commonsense knowledge from the neural commonsense knowledge model (i.e., source model) to the pre-trained model (i.e., target model) so that it can generalize better to downstream tasks requiring commonsense reasoning ability. 

Compared to conventional knowledge transfer methods such as knowledge distillation, \method faces a unique challenge: the source model and the target model are heterogeneous because they are trained on different data with different objectives. As such, we can not simply feed a batch of data to both of the models and train the target model to match the source model's logits or feature distribution. To alleviate this problem, we propose a two-stage knowledge transfer scheme as illustrated in Figure \ref{fig:framework}. To be specific, we first use natural texts to form queries for retrieving commonsense knowledge (in text form) from the neural commonsense knowledge model. We then construct training data with two novel commonsense-related self-supervised objectives based on the retrieved commonsense knowledge and the corresponding natural text. Finally, we train the target model on the constructed training data to inject commonsense knowledge retrieved from the source model. We describe our method to extract commonsense knowledge from a neural commonsense knowledge model and the proposed commonsense-related self-supervised objectives in detail in this section. 
\subsection{Commonsense Knowledge Extraction}
We first describe the source model, i.e., neural commonsense knowledge model, in the \method framework. It is a transformer~\citep{DBLP:conf/nips/VaswaniSPUJGKP17} language model trained on commonsense knowledge graphs like \textsc{Atomic}~\citep{DBLP:conf/aaai/SapBABLRRSC19} and ConceptNet~\citep{DBLP:conf/aaai/SpeerCH17} with the objective of predicting the object (i.e., commonsense inference) with the subject (i.e., natural text) and relation as input. For example, given a commonsense tuple (s=“take a nap", r=Causes, o=“have energy"), the neural commonsense knowledge model is trained to generate $o$ given $s$ and $r$ as inputs. After training, it can generate accurate, representative knowledge for new, unseen entities and events.

To extract commonsense knowledge stored in a neural commonsense knowledge model, we use a natural sentence as the subject $s$ (e.g., he wants to cook a meal) and concatenate it with a randomly selected commonsense relation $r$ (e.g., xNeed) from a pre-defined set to form a prompt (e.g., he wants to cook a meal xNeed ). We then feed the prompt to the neural commonsense knowledge model and use it to generate a commonsense inference (e.g., to buy ingredients). In this way, the commonsense knowledge generation process resembles the way in which the neural commonsense knowledge model is trained. As such, we can get commonsense inferences of relatively high qualities.

Using a neural commonsense knowledge model as a knowledge source has two advantages. On one hand, compared to the previous method~\citep{DBLP:journals/corr/abs-1909-09743} using a symbolic commonsense knowledge graph, a neural commonsense knowledge model can generalize to unseen subjects, thus enabling us to refine the target pre-trained model on large-scale natural text corpus together with its commonsense inferences. As such, the resulting model can better adapt to downstream tasks which are formulated in diverse natural texts. On the other hand, compared to another method~\citep{DBLP:conf/iclr/ZhouLSL021} that only uses plain text and is thus limited to the surface form of naturally occurring text, the use of a neural commonsense knowledge model provides much denser commonsense knowledge including a diverse set of commonsense relations between natural texts and the underlying commonsense knowledge.


\subsection{Commonsense Knowledge Injection} 

After commonsense knowledge extraction, we need to inject the extracted commonsense knowledge into the target model. A straightforward solution is to use sequence-level knowledge distillation~\citep{DBLP:conf/emnlp/KimR16} and continually train the student to generate retrieved commonsense inference given the original text and commonsense relation. However, this can be sub-optimal due to the domain discrepancy between commonsense knowledge and natural text, which introduces the catastrophic forgetting problem~\citep{Kirkpatrick3521} and hurts the performance on downstream tasks, which is also recently confirmed by~\citet{DBLP:journals/corr/abs-2109-02572}.

To better inject the extracted commonsense knowledge into a pre-trained model without suffering from catastrophic forgetting so that its capability on general NLP tasks is retained (or even improved), we propose two commonsense-related self-supervised objectives: \textit{commonsense text infilling} and \textit{commonsense relation prediction}. The former objective is generative while the latter is a discriminative objective. We refine the pre-trained model by multi-tasking on both objectives so that the model can better adapt to tasks requiring either generative or discriminative commonsense reasoning ability.

\begin{figure}[t]
\centering
\small
     \includegraphics[width=\linewidth]{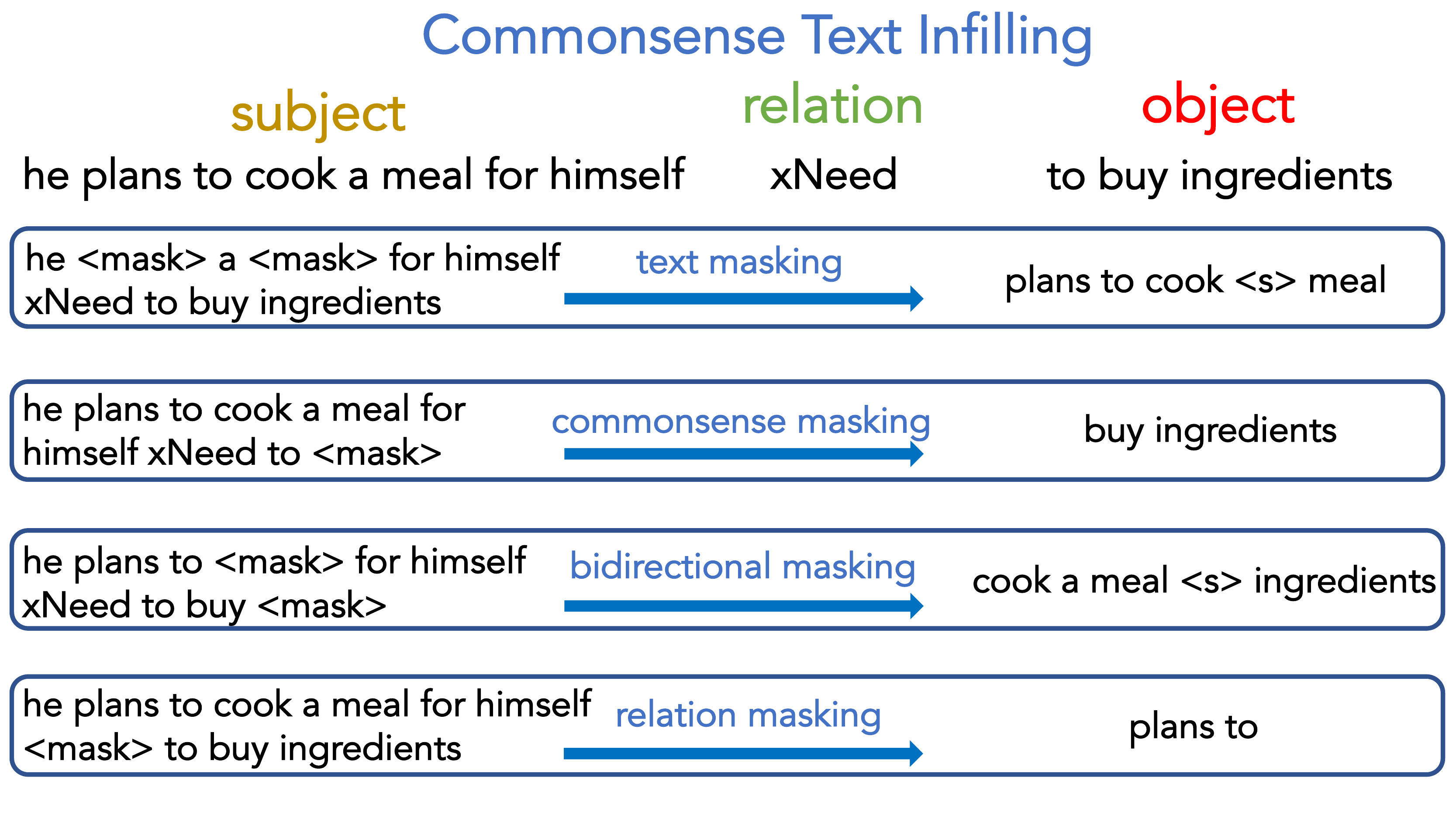}
     \caption{Illustration of the commonsense text infilling objective. Given a commonsense tuple constructed in the commonsense knowledge retrieval phase, we randomly mask text spans in the commonsense tuple following different patterns and train the pre-trained model to reconstruct the masked spans. \ronan{Should the <mask> be ``xNeed''?}} 
 \label{fig:csti}
 \vspace{-0.5cm}
 \end{figure}

\paragraph{Commonsense Text Infilling}

Commonsense text infilling is a simple extension to the conventional text infilling objective used for pre-training BART and T5. It transforms each sentence to a commonsense tuple similar to that in a commonsense knowledge graph by appending the commonsense relation and the generated commonsense inference. We then mask text spans in the commonsense tuple by randomly selecting one masking scheme among \textit{text masking}, \textit{commonsense masking}, \textit{bidirectional masking}, and \textit{relation masking}. As illustrated in Fig 2, these masking strategies selectively mask different components in the input commonsense tuple and lead to different optimization objectives. Specifically, these masking schemes mask either spans in natural text ($\text{P}(s|\Tilde{s},r,o)$), commonsense inference ($\text{P}(o|s,r,\Tilde{o})$), natural text/commonsense inference ($\text{P}(s,o|\Tilde{s},r,\Tilde{o})$), or commonsense relation ($\text{P}(r|s,\Tilde{r},o)$), respectively. We then train the model to predict the masked spans autoregressively. The diverse masking strategies provide more diverse training signals compared to random masking, thus enabling the model to better align the surface form of human language and the underlying commonsense knowledge.

In addition, unlike the conventional practice in masked span infilling objective that randomly masks text spans with the same probability, we propose to mask text spans including concepts (tokens recognized as nouns or verbs by a Spacy POS tagger) with a higher probability so that the model will be trained to predict concepts more frequently compared to non-content words that are generally not related to commonsense reasoning.

\begin{figure}[t]
\centering
\small
     \includegraphics[width=\linewidth]{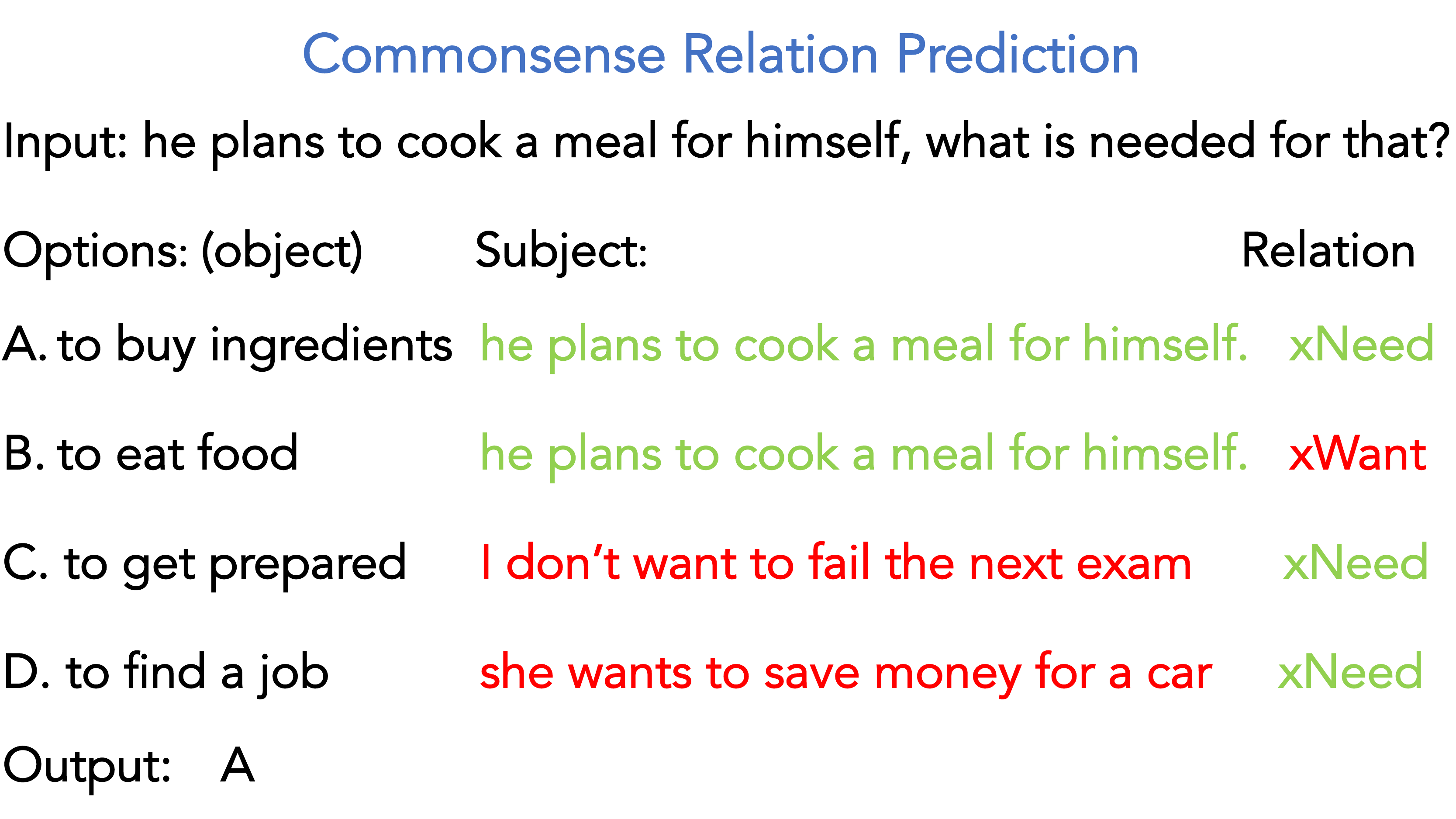}
     \caption{Illustration of the commonsense relation prediction objective.  We train the pre-trained model to predict the correct commonsense inference given the subject and relation from three distractors generated with either different subjects or relations as inputs.}
 \label{fig:csrp}
 \vspace{-0.5cm}
 \end{figure}

\begin{table*}[!h]
\centering
	\centering
	\resizebox{\textwidth}{!}{
		\begin{tabular}{lcccccc}
			\toprule
			\textbf{Methods} &
            \textbf{CSQA} &  \textbf{OBQA} & \textbf{PIQA} & \textbf{aNLI} & \textbf{\textsc{SocialIQA}} & \textbf{COPA} \\
			\midrule
			BERT-base & 53.08($\pm$0.16) & 57.60($\pm$0.8) & 64.86($\pm$0.52) & 61.88($\pm$0.56) & 64.3($\pm$0.4) & 67.3($\pm$0.4)  \\ 
			ERNIE-base & 54.06($\pm$0.12) & 58.90($\pm$0.9) & 66.47($\pm$0.58) & 63.04($\pm$0.46) & 65.1($\pm$0.4) & 68.9($\pm$0.4)   \\ 
			KnowBERT & 53.88($\pm$0.15) & 58.50($\pm$0.8) & 66.61($\pm$0.63) & 63.18($\pm$0.52)  & 65.4($\pm$0.5) & 69.4 ($\pm$0.4)  \\ 
			COMET & 45.32($\pm$0.28) & 51.20($\pm$1.1) & 60.73($\pm$0.51) & 57.63($\pm$0.61)  & 60.2($\pm$0.7) & 69.1 ($\pm$0.5)  \\ 
			 \midrule
			 T5-base & 61.88($\pm$0.08) & 58.20($\pm$1.0) & 68.14($\pm$0.73) & 61.10($\pm$0.38)  & 65.1($\pm$0.5) & 71.4 ($\pm$0.7)   \\
			 T5-base + TI & 62.05($\pm$0.17) & 58.43($\pm$0.8) & 68.32($\pm$0.66) & 61.42($\pm$0.32) & 65.3($\pm$0.4) & 71.8 ($\pm$0.8)   \\ 
			 T5-base + SSM & 62.37($\pm$0.25) & 58.60($\pm$0.9) & 68.48($\pm$0.65) & 61.57($\pm$0.44) & 65.5($\pm$0.5) & 72.1 ($\pm$0.6)  \\ 
			 T5-base + KD & 61.83($\pm$0.42) & 56.54($\pm$0.7) & 67.35($\pm$0.63) & 60.94($\pm$0.66)  & 64.8($\pm$0.5)  & 71.0 ($\pm$1.0)  \\ 
			 T5-base + CSKG (TI) & 60.22($\pm$0.40) & 56.17($\pm$0.8) & 66.51($\pm$0.57) &  59.92($\pm$0.47)  & 62.7($\pm$0.7) & 68.5 ($\pm$1.1)  \\
			 T5-base + CSKG (Rule) & 63.10($\pm$0.35) & 57.97($\pm$0.8) & 68.27($\pm$0.71) & 60.15($\pm$0.51)  & 65.7($\pm$0.4)  & \underline{72.4 ($\pm$0.9)}  \\ 
			 CALM  & \underline{63.32($\pm$0.35)} & \underline{60.90($\pm$0.4)} & \underline{71.01($\pm$0.61)} & \underline{63.20($\pm$0.52)} & \underline{66.0($\pm$0.5)} & 72.2 ($\pm$0.8) \\
			\midrule
            \baby-base  & \bf 64.11($\pm$0.31) & \bf 61.58($\pm$0.5) & \bf 72.26($\pm$0.61) & \bf 64.37($\pm$0.49) & \bf 67.3($\pm$0.4) & \bf 73.4 ($\pm$0.5)  \\ 
            \baby w/ GPT-2 & 60.39($\pm$0.61) & 56.95($\pm$0.7) & 68.48($\pm$0.44) &  60.14($\pm$0.52)  & 66.2($\pm$0.6) & 72.8 ($\pm$1.0)  \\
			\bottomrule
	\end{tabular}
	}
	\caption{\textbf{Experimental results on base-size models.} Best models are bold and second best ones are \underline{underlined} within each metric. Mean and standard deviation of 3 different runs with different random seeds are reported.} 
	\vspace{-0.4cm}
	\label{tab:result-base}
\end{table*}

\paragraph{Commonsense Relation Prediction}

While the commonsense text infilling objective encourages the pre-trained model to align natural texts and their commonsense inferences, it is always trained on \textit{valid} commonsense tuples.
This can be sub-optimal because we also want the model to be capable of discriminating invalid commonsense inferences, which is important for many commonsense-related downstream tasks. 

To this end, we introduce a commonsense relation prediction task that trains the model to distinguish the correct commonsense inference corresponding to the input sentence and the commonsense relation from distractors. To be specific, the commonsense relation prediction objective is formulated as a multi-choice QA problem with an input sentence as the context, a commonsense relation as the question, and a set of four commonsense inferences as options. The set of options consists of one correct commonsense inference, which is generated by the neural commonsense model with the input sentence and commonsense relation as input, and three carefully curated distractors (i.e., negative examples) generated by the same neural commonsense knowledge model with different inputs. As illustrated in Figure 3, among the three distractors, one is generated with an input composed of the same sentence and a different commonsense relation, and another two are generated with an input composed of different sentences with the same commonsense relation. In this way, the model learns to align the natural texts with valid commonsense knowledge while also distinguishing commonsense inferences that do not make sense. Moreover, this objective is formulated as a multi-choice QA task that closely resembles several downstream commonsense-related tasks such as CommonsenseQA and \textsc{SocialIQA}, thus enabling easier transfer especially when labeled training examples are scarce.


\section{Experiments}


\subsection{Experimental Settings}

\paratitle{Models} In our experiments we apply \method to refine T5~\citep{raffel2019exploring}, a popular model pre-trained with the text infilling objective. We experiment with both T5-base and T5-large, which consist of 220 million and 774 million parameters respectively, as the target model in the \method framework. We do not experiment with extremely large models like T5-11B because of the resource constraints and the fact that these models are hard to deploy in real-world applications. We use COMET-\textsc{Atomic}$_{20}^{20}$, a state-of-the-art neural commonsense knowledge model that can generate accurate, representative knowledge for new, unseen entities and events, as the source model. It is initialized with BART and continually trained on \textsc{Atomic}$_{20}^{20}$~\citep{DBLP:conf/aaai/HwangBBDSBC21}, a new general purpose commonsense knowledge graph. 

\paratitle{Data} We randomly sample a subset consisting of 10 million sentences from the English Wikipedia and the BookCorpus~\citep{DBLP:conf/iccv/ZhuKZSUTF15}, which is used for pre-training BERT and its variants. We select a set of representative commonsense relations including intent, reason, effect, need, want, and react from relations used to train COMET-\textsc{Atomic}$_{20}^{20}$. For each sentence, we randomly sample two relations and retrieve the corresponding commonsense explanation from COMET$_{20}^{20}$. We randomly select one relation-explanation pair to form the input example and leave another as the distractor for the commonsense relation prediction objective. 

\begin{table*}[!h]
\centering
	\centering
	\resizebox{\textwidth}{!}{
		\begin{tabular}{lcccccc}
			\toprule
			\textbf{Methods} &
            \textbf{CSQA} &  \textbf{OBQA} & \textbf{PIQA} & \textbf{aNLI} & \textbf{\textsc{SocialIQA}} & \textbf{COPA} \\
			\midrule
			T5-large & 69.81($\pm$1.02) & 61.40($\pm$1.0) & 72.19($\pm$1.09) & 75.54($\pm$1.22) & 71.3($\pm$0.8) & 83.6($\pm$1.1) \\
			CALM-large  & \underline{71.31($\pm$0.04)} & \underline{66.00($\pm$1.0)} & \underline{75.11($\pm$1.65)} & \underline{77.12($\pm$0.34)} & \underline{72.7($\pm$0.7)} & \underline{84.9($\pm$1.0)} \\
			\midrule
            \baby-large  & \bf 72.15($\pm$0.61) & \bf 66.70($\pm$1.1) & \bf 76.07($\pm$0.95) & \bf {77.94($\pm$0.59)} & \bf 73.8($\pm$0.8) & \bf {86.0($\pm$1.2) } \\ 
			\bottomrule
	\end{tabular}
	}
	\caption{\textbf{Experimental results on large-size models.} Best models are bold and second best ones are \underline{underlined} within each metric. Mean and variance of 3 different runs with different random seeds are reported.}
	\vspace{-0.5cm}
	\label{tab:result-large}
\end{table*}

\paratitle{Training} We refine the pre-trained models on the self-supervised examples constructed with the sampled 10 million sentences for 100k steps with a batch size of 1024, a maximum sequence length of 256, and a learning rate of 5e-5/2e-5 for base-size and large-size models respectively with a linear warm-up for the first 8,000 updates. After knowledge transfer, we fine-tune the models on downstream tasks by formulating the tasks into text-to-text problems. Pre-training and fine-tuning details are included in the Appendix.

\paratitle{Evaluation} We evaluate the continual pre-trained models on downstream tasks that require commonsense reasoning including \textbf{CommonsenseQA}~\citep{talmor2018commonsenseqa}, \textbf{OpenbookQA}~\citep{mihaylov2018can}, \textbf{PIQA}~\citep{Bisk2020}, \textbf{aNLI}~\citep{bhagavatula2019abductive}, \textbf{COPA}~\citep{DBLP:conf/aaaiss/RoemmeleBG11},
and \textbf{\textsc{SocailIQA}}~\citep{sap2019socialiqa}
In addition to the conventional fully supervised setting, we also test our approach in the few-shot setting by varying the percentage of labeled examples from the original training set used for fine-tuning. The idea is that limited labeled examples can only help the model understand the task but are insufficient for the model to acquire enough commonsense knowledge to solve the task. As such, it requires the model to store enough commonsense knowledge in its parameters to succeed in the few-shot setting. For both the settings, we report the results on the official development set and tune the hyperparameters based on the models’ performance on an in-house split dev set. We report the mean and variance of 3 individual runs with different random seeds because most datasets are relatively small, which makes the variance in results non-negligible.

\paratitle{Baselines} We compare our approach with methods that continually train a pre-trained model with different objectives. We divide the baselines into two categories based on the source of their supervision. The first category includes methods that only exploit general text corpus, including (1) \textbf{T5 + TI} that continually pre-trains the public checkpoint of T5 with the same text infilling objective for more steps, (2) \textbf{T5 + SSM} that also continual pre-trains T5 with the text infilling objective, but use salient span masking~\citep{DBLP:conf/emnlp/RobertsRS20} instead of random masking for data construction, (3) \textbf{(T5 + KD)} that uses sequence-level knowledge distillation~\citep{DBLP:conf/emnlp/KimR16} for knowledge transfer, where the student model is trained with the teacher output (i.e., $\text{P}(o|s,r)$), and (4) \textbf{CALM}~\citep{DBLP:conf/iclr/ZhouLSL021} that uses novel self-supervised objectives to construct concept-centric self-supervision from general text corpus. The second category instead exploits CSKG, including (5) \textbf{T5 + CSKG (TI)} train T5 with the text infilling objective on tuples in a CSKG, and (6) \textbf{T5 + CSKG (Rule)}~\citep{DBLP:journals/corr/abs-1909-09743} that uses manually defined rules to construct training examples from a CSKG. We also include a \textbf{COMET} baseline where we directly fine-tune the pre-trained COMET-\textsc{Atomic}$_{20}^{20}$ model for the downstream tasks to verify the necessity of \method, and a \textbf{CKT w/ GPT-2} baseline where the commonsense inferences are generated by a pre-trained GPT-2 large model to verify whether the gain comes from transferring the commonsense knowledge from COMET, or simply from data augmentation from another generative model. For a fair comparison, we use the same data and training steps compared to our approach for baselines from the first category and use \textsc{Atomic}$_{20}^{20}$, on which the teacher model in our framework is pre-train on, as the commonsense knowledge graph. For reference, we also include some popular knowledge-enhanced pre-trained models including ERNIE~\citep{zhang2019ernie} and KnowBERT~\citep{peters2019knowledge}. 

\begin{figure}
\centering
\begin{minipage}{.25\textwidth}
  \centering
  \includegraphics[width=\textwidth]{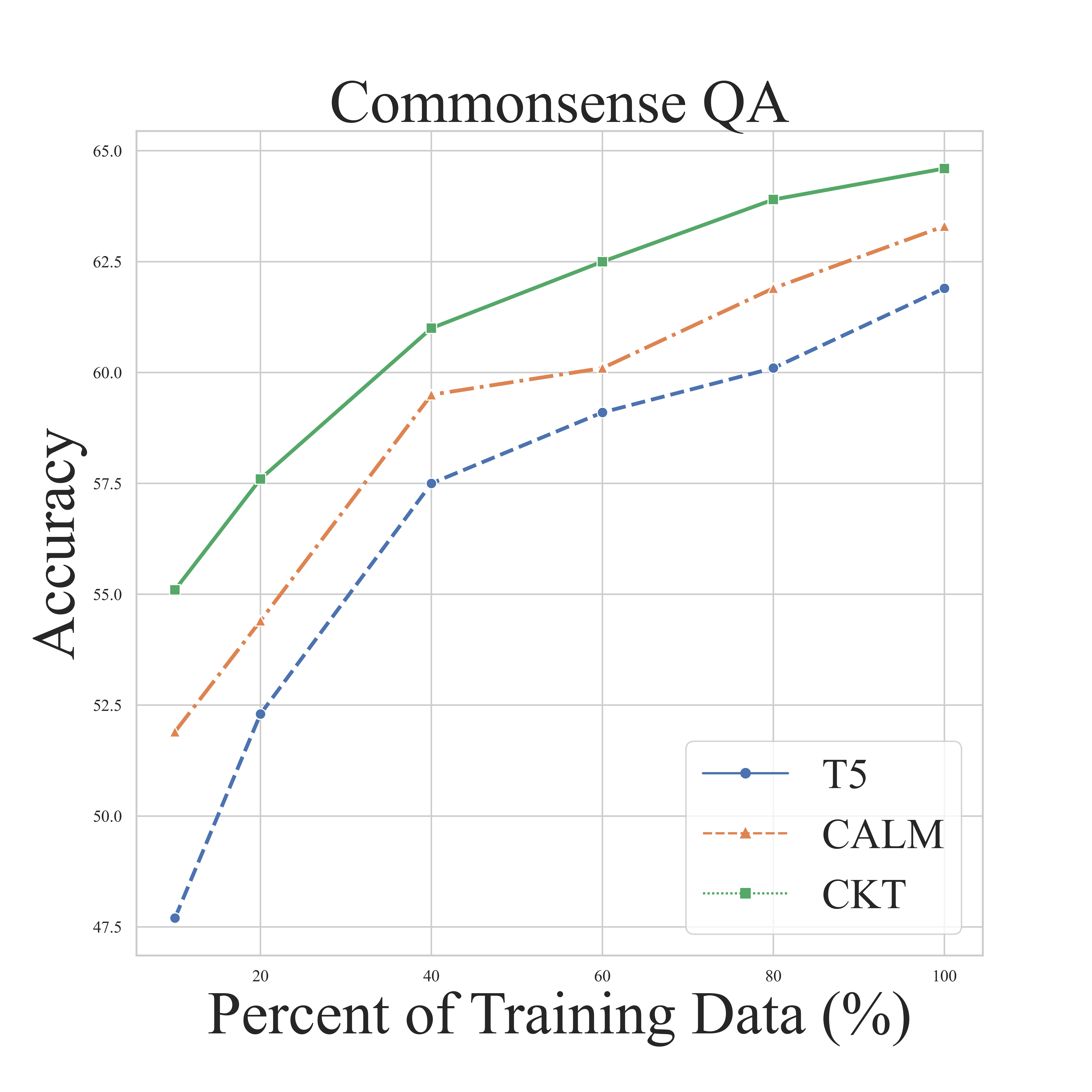}
  \label{fig:test1}
\end{minipage}%
\begin{minipage}{.25\textwidth}
  \centering
  \includegraphics[width=\textwidth]{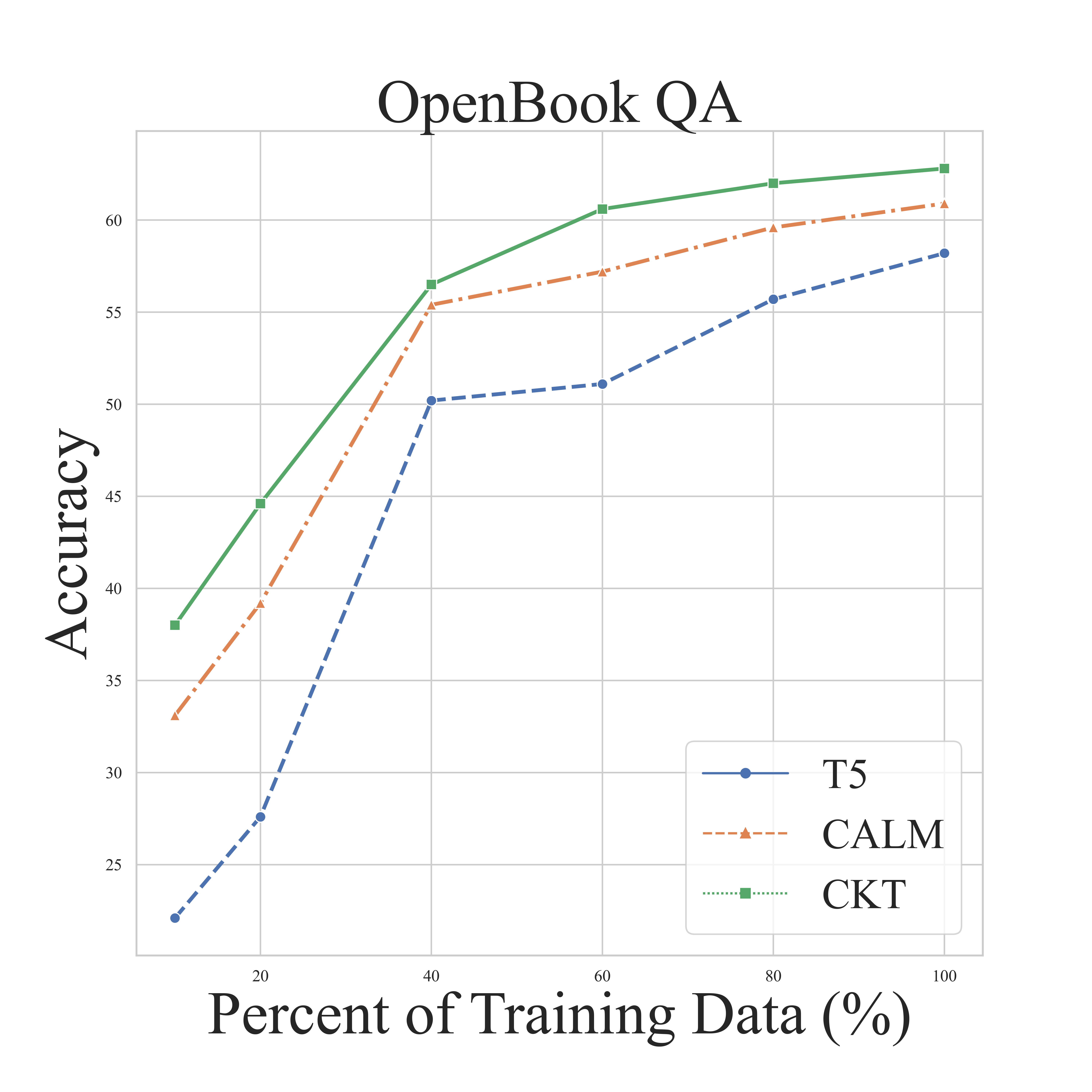}
  \label{fig:test2}
\end{minipage}
\vspace{-0.5cm}
\caption{Performance of compared base-size models fine-tuned with
different fraction of the datasets.}
\vspace{-0.4cm}
\label{fig:fewshot}
\end{figure}

\subsection{Fully-supervised Results}

We first present results in the fully-supervised setting. Results on base-size models are presented in Table \ref{tab:result-base}. We can see that our approach yields significant improvement compared to the T5 baseline (up to 4 absolute scores) and consistently outperform CALM, the state-of-the-art method of injecting commonsense knowledge into PTLMs.

\begin{table*}[]
\centering
	\centering
	\resizebox{0.85\textwidth}{!}{
		\begin{tabular}{lcccccc}
			\toprule
			\textbf{Methods} &
            \textbf{CSQA} &  \textbf{OBQA} & \textbf{PIQA} & \textbf{aNLI} & \textbf{\textsc{SIQA}} & \textbf{COPA} \\
            \midrule
			 T5-base & 61.88 & 58.20 & 68.14 & 61.10  & 65.1 & 71.4   \\
            \baby-base  &  64.57 &  62.77 &  73.26 & 64.75 &  68.3 &  73.4 \\
            \midrule
			\multicolumn{7}{c}{\textit{Objective Analysis}} \\
			\midrule
            \baby-base w/o CSTI  & 62.58 & 60.97 & 70.61 & 62.11 & 66.5 & 72.0 \\
            \baby-base w/o text masking  & 62.98 & 61.74 & 72.55 & 63.81 & 67.7 & 72.8\\
            \baby-base w/o commonsense masking   & 63.61 & 62.03 & 72.83 & 64.40 & 67.5 & 72.7 \\
            \baby-base w/o bidirectional masking  & 63.52 & 62.11 & 72.30 & 64.24 & 67.6 & 72.9 \\
            \baby-base w/o relation masking   & 64.12 & 62.48 & 73.31 & 64.57 & 67.4 & 72.7 \\
            \baby-base w/o CSRP  & 63.12 & 62.07 & 72.44 & 64.11 & 67.5 & 72.6 \\
            \baby-base w/ random distractors   & 64.04 & 62.29 & 72.95 & 64.48 & 68.0 & 73.1 \\
            \midrule
			\multicolumn{7}{c}{\textit{Multi-task versus  Sequential Transfer}} \\
			\midrule
            \baby-base (CSTI $\rightarrow$ CSRP)   & 64.69 & 62.51 & 73.35 & 64.11 & 67.9 & 73.5 \\
            \baby-base (CSRP $\rightarrow$ CSTI)   & 63.49 & 61.33 & 71.54 & 63.41 & 67.0 & 72.0 \\
            \midrule
			\multicolumn{7}{c}{\textit{Corpus Size}} \\
			\midrule
            \baby-base w/ $10\%$ data   & 64.18 & 62.21 & 71.86 & 64.31 & 67.7 & 73.1 \\
            \baby-base w/ $50\%$ data    & 64.45 & 62.66 & 73.10 & 64.72 & 68.2 & 73.4 \\
			\bottomrule
	\end{tabular}
	}
	\caption{Analysis of the proposed commonsense knowledge transfer framework. CSTI and CSRP denote the commonsense text infilling objective and the commonsense relation prediction objective, respectively.}
	\label{tab:result-analysis}
	\vspace{-0.3cm}
\end{table*}

In addition, we observe that simply using continual training with the original text-infilling objective or its variant with salient span masking only marginally improves the performance. Surprisingly, training with text infilling on a commonsense knowledge graph leads to degraded performance compared to the T5 baseline. We suspect this is because the commonsense tuples in commonsense knowledge graphs are generally too short and simple, making the pre-trained model unable to reason within relatively long contexts which is crucial for most downstream tasks. Moreover, we find that continually pre-training with training data constructed with commonsense tuples in a commonsense knowledge graph following manually designed rules leads to improvements in certain tasks. However, the improvement is inconsistent across different tasks and it even hurts the performance on certain tasks, which may be because the rules for constructing training data are tailored for certain tasks like CSQA. The inferior performance of using commonsense knowledge graphs as data sources also confirms the need of using natural text corpus during continual pre-training for better adapting to diverse downstream tasks. We also find directly applying sequence-level KD and training the student to mimic the teacher on the commonsense tuple generation task fails to improve the performance because the task is not general enough and thus cannot transfer to diverse downstream tasks well. Moreover, directly fine-tuning COMET or using GPT-2 as the commonsense knowledge source results in very poor performance. This confirms the necessity of \method and shows that it is actually transferring commonsense knowledge instead of simple text augmentation.

To further confirm the effectiveness of \method, we apply it to T5-large and compare it to competitive baselines in the base-size experiments. The results are presented in Table \ref{tab:result-large}. We can see that our approach consistently outperforms T5-large and CALM-large. This suggests that our approach can successfully generalize to large-size pre-trained models.        

\subsection{Few-shot Results}

Injecting commonsense knowledge into pre-trained models is important because it enables the model to reason and generalize to unseen examples while observing only a few labeled examples. To this end, we fine-tune the compared models with different fractions of labeled training data to investigate the transition of the behavior of our model and baselines from the low-resource regime to the fully-supervised setting (Fig. \ref{fig:fewshot}). 
We observe that the performance improvement of our approach compared to the baselines is more significant in
the low-resource regime. This shows
that \method can successfully transfer commonsense knowledge into pre-trained models so that they can generalize well while seeing only a small part of training data. This may also help the model reduce the risk/tendency of fitting the spurious correlations in the annotated datasets and thus generalize better.

\subsection{Analysis}

To better understand the proposed \method framework and the role of its different components, we conduct an ablation study about the impact of different proposed objectives, the impact of multi-tasking the commonsense-related self-supervised objective versus sequentially training, and the impact of the size of natural text corpus used for transfer (see Table \ref{tab:result-analysis}). 

\paratitle{Impact of Objectives} We find that both the proposed objectives contribute to the performance improvement of our approach. The commonsense text infilling objective is shown to be more critical than the commonsense relation prediction task. We suspect this is because commonsense text infilling resembles the vanilla text infilling objective with which the T5 models are pre-trained, thus preventing the model from catastrophic forgetting. In addition, all four masking strategies are beneficial, and their contribution varies for different downstream tasks. This confirms the necessity of a diverse masking scheme. Moreover, our strategy for constructing distractors outperforms the random counterpart, demonstrating the necessity of hard negative examples for the commonsense relation prediction task.

\paratitle{Multi-task versus Sequential Transfer} As for the training order between the two objectives, we find that starting from the commonsense text infilling task and then switching to the commonsense relation prediction task performs similarly with our multi-tasking strategy while significantly outperforming its counterpart training with the reverse direction. We think this is because the commonsense text infilling objective resembles the original pre-training while the commonsense relation prediction is more similar to downstream tasks. We opt for the multi-tasking strategy for simplicity.

\paratitle{Impact of Corpus Size} We find that \method significantly outperforms both the T5 baseline and the competitive CALM method with only 10 percent of the full data used for distillation. Nevertheless, the performance improvement also confirms that our approach can benefit from the accessibility of large-scale natural texts. For base-size models, the performance improvements seem to saturate after 10 million sentence pairs. However, we anticipate that larger-size models may still benefit from a larger amount of data, and leave this for future work.

\section{Related Work}
\label{sec:related_work}


\paratitle{Knowledge-augmented Pre-trained Models} 
A number of recent works have examined the problem of incorporating world knowledge with the pre-trained models. A number of works use an external knowledge base to incorporate entity knowledge with pre-trained models~\cite{zhang2019ernie,peters2019knowledge,wang2020k,liu2020kgbart}. However, these approaches require specialized resources like knowledge bases which are non-trivial to seek, thus limiting the domain they can be applied to. \citet{Xiong2020Pretrained} proposed a novel entity replacement detection objective that incorporates Wikipedia to encode world knowledge into a BERT-like pre-trained model. \citet{DBLP:journals/corr/abs-2012-03551} proposed a generative and discriminative framework that pre-trains the model to complete and correct knowledge spans. The aforementioned approaches generally focus on factual knowledge of entities while our work mainly focuses on commonsense knowledge. 

\paratitle{Commonsense Reasoning for NLP}
Several recent studies~\citep{talmor2018commonsenseqa, sap2019socialiqa,zhou2020evaluating, DBLP:conf/emnlp/LinZSZBCR20,DBLP:conf/naacl/XuZGXMW21} evaluate the performance of several pre-trained language models on tasks that require commonsense reasoning and find that it is still very hard for pre-trained language models to match or exceed human-level performance. Therefore, approaches to improve the commonsense reasoning ability of pre-trained language models has attracted much attention. These approaches can be divided into two categories. The first category focuses on incorporating an external commonsense knowledge graph for commonsense reasoning. For example,~\citet{DBLP:conf/emnlp/LinCCR19},~\citet{DBLP:journals/corr/abs-2109-02572}, and~\citet{DBLP:conf/aaai/LiuW0PY21} propose to exploit structured symbolic commonsense knowledge graphs to perform commonsense reasoning. 
The second one instead attempts to inject commonsense knowledge into the parameters of pre-trained models. For example,~\citet{DBLP:journals/corr/abs-1908-06725, DBLP:journals/corr/abs-1909-09743} proposed to use manually designed rules to construct commonsense related training examples from commonsense knowledge graphs. \citet{DBLP:conf/iclr/ZhouLSL021} instead only relies on general text corpus and proposed two concept-centric self-supervised objectives to refine pre-trained models with commonsense knowledge. 

\section{Conclusion}

We introduce commonsense knowledge transfer, a framework to transfer the commonsense knowledge stored in a neural commonsense knowledge model to a general-purpose pre-trained model. Our method extracts commonsense knowledge from the source model to construct self-supervised training data for the target model. Empirical results show that our approach consistently outperforms previous methods for improving the commonsense reasoning ability of pre-trained models that exploit either symbolic knowledge graphs or texts alone. 

\section*{Limitations}

In our experiments, we use T5-base and T5-large models as the target model since they are widely-used, representative pre-trained seq2seq models and use COMET-\textsc{Atomic}$_{20}^{20}$ as the commonsense knowledge source. However, there are other pre-trained seq2seq models such as BART, and neural commonsense models such as COMET that we did not experiment with. Moreover, we only experimented with 10 million randomly sampled sentences from the English Wiki and BookCorpus datasets. It would be interesting to investigate whether continually pre-training with a larger scale dataset can further improve the performance. 

\section*{Ethical Considerations}

Our work focuses on improving the commonsense reasoning ability of pre-trained language models. It probably does not introduce extra ethical concerns. However, in commonsense knowledge extraction, the neural commonsense knowledge model may generate unexpected (e.g., biased) commonsense inferences, and training with these inferences may lead to additional bias in the pre-trained model. Nevertheless, all pre-trained language models contain bias and should be examined.


\bibliographystyle{acl_natbib}
\bibliography{anthology,acl2021}

\begin{thebibliography}{53}
\expandafter\ifx\csname natexlab\endcsname\relax\def\natexlab#1{#1}\fi

\bibitem[{Bender and Koller(2020)}]{DBLP:conf/acl/BenderK20}
Emily~M. Bender and Alexander Koller. 2020.
\newblock \href {https://doi.org/10.18653/v1/2020.acl-main.463} {Climbing
  towards {NLU:} on meaning, form, and understanding in the age of data}.
\newblock In \emph{Proceedings of the 58th Annual Meeting of the Association
  for Computational Linguistics, {ACL} 2020, Online, July 5-10, 2020}, pages
  5185--5198. Association for Computational Linguistics.

\bibitem[{Bhagavatula et~al.(2020)Bhagavatula, Bras, Malaviya, Sakaguchi,
  Holtzman, Rashkin, Downey, Yih, and Choi}]{bhagavatula2019abductive}
Chandra Bhagavatula, Ronan~Le Bras, Chaitanya Malaviya, Keisuke Sakaguchi, Ari
  Holtzman, Hannah Rashkin, Doug Downey, Scott Wen-tau Yih, and Yejin Choi.
  2020.
\newblock Abductive commonsense reasoning.
\newblock In \emph{ICLR}.

\bibitem[{Bisk et~al.(2020)Bisk, Zellers, Bras, Gao, and Choi}]{Bisk2020}
Yonatan Bisk, Rowan Zellers, Ronan~Le Bras, Jianfeng Gao, and Yejin Choi. 2020.
\newblock Piqa: Reasoning about physical commonsense in natural language.
\newblock In \emph{Thirty-Fourth AAAI Conference on Artificial Intelligence
  (AAAI)}.

\bibitem[{Bosselut et~al.(2019)Bosselut, Rashkin, Sap, Malaviya, Celikyilmaz,
  and Choi}]{DBLP:conf/acl/BosselutRSMCC19}
Antoine Bosselut, Hannah Rashkin, Maarten Sap, Chaitanya Malaviya, Asli
  Celikyilmaz, and Yejin Choi. 2019.
\newblock {COMET:} commonsense transformers for automatic knowledge graph
  construction.
\newblock In \emph{{ACL} {(1)}}, pages 4762--4779. Association for
  Computational Linguistics.

\bibitem[{Conneau and Kiela(2018)}]{senteval}
Alexis Conneau and Douwe Kiela. 2018.
\newblock Senteval: An evaluation toolkit for universal sentence
  representations.
\newblock In \emph{{LREC}}.

\bibitem[{Cui and Chen(2021)}]{DBLP:journals/corr/abs-2109-02572}
Wanyun Cui and Xingran Chen. 2021.
\newblock \href {http://arxiv.org/abs/2109.02572} {Enhancing language models
  with plug-and-play large-scale commonsense}.
\newblock \emph{CoRR}, abs/2109.02572.

\bibitem[{Devlin et~al.(2019)Devlin, Chang, Lee, and
  Toutanova}]{DBLP:conf/naacl/DevlinCLT19}
Jacob Devlin, Ming{-}Wei Chang, Kenton Lee, and Kristina Toutanova. 2019.
\newblock {BERT:} pre-training of deep bidirectional transformers for language
  understanding.
\newblock In \emph{{NAACL-HLT} {(1)}}, pages 4171--4186. Association for
  Computational Linguistics.

\bibitem[{Dolan and Brockett(2005)}]{mrpc}
William~B. Dolan and Chris Brockett. 2005.
\newblock Automatically constructing a corpus of sentential paraphrases.
\newblock In \emph{IWP@IJCNLP}.

\bibitem[{Falcon(2019)}]{falcon2019pytorch}
WA~Falcon. 2019.
\newblock Pytorch lightning.
\newblock \emph{GitHub. Note:
  https://github.com/PyTorchLightning/pytorch-lightning}, 3.

\bibitem[{He et~al.(2020)He, Jiang, Xiao, and
  Liu}]{DBLP:journals/corr/abs-2012-03551}
Bin He, Xin Jiang, Jinghui Xiao, and Qun Liu. 2020.
\newblock \href {http://arxiv.org/abs/2012.03551} {Kgplm: Knowledge-guided
  language model pre-training via generative and discriminative learning}.
\newblock \emph{CoRR}, abs/2012.03551.

\bibitem[{Hewitt and Manning(2019)}]{DBLP:conf/naacl/HewittM19}
John Hewitt and Christopher~D. Manning. 2019.
\newblock \href {https://doi.org/10.18653/v1/n19-1419} {A structural probe for
  finding syntax in word representations}.
\newblock In \emph{Proceedings of the 2019 Conference of the North American
  Chapter of the Association for Computational Linguistics: Human Language
  Technologies, {NAACL-HLT} 2019, Minneapolis, MN, USA, June 2-7, 2019, Volume
  1 (Long and Short Papers)}, pages 4129--4138. Association for Computational
  Linguistics.

\bibitem[{Hinton et~al.(2015)Hinton, Vinyals, and
  Dean}]{DBLP:journals/corr/HintonVD15}
Geoffrey~E. Hinton, Oriol Vinyals, and Jeffrey Dean. 2015.
\newblock \href {http://arxiv.org/abs/1503.02531} {Distilling the knowledge in
  a neural network}.
\newblock \emph{CoRR}, abs/1503.02531.

\bibitem[{Hosseini et~al.(2021)Hosseini, Broniatowski, and
  Diab}]{DBLP:journals/corr/abs-2112-08615}
Pedram Hosseini, David~A. Broniatowski, and Mona~T. Diab. 2021.
\newblock Commonsense knowledge-augmented pretrained language models for causal
  reasoning classification.
\newblock \emph{CoRR}, abs/2112.08615.

\bibitem[{Hwang et~al.(2021)Hwang, Bhagavatula, Bras, Da, Sakaguchi, Bosselut,
  and Choi}]{DBLP:conf/aaai/HwangBBDSBC21}
Jena~D. Hwang, Chandra Bhagavatula, Ronan~Le Bras, Jeff Da, Keisuke Sakaguchi,
  Antoine Bosselut, and Yejin Choi. 2021.
\newblock (comet-) atomic 2020: On symbolic and neural commonsense knowledge
  graphs.
\newblock In \emph{{AAAI}}, pages 6384--6392. {AAAI} Press.

\bibitem[{Kim and Rush(2016)}]{DBLP:conf/emnlp/KimR16}
Yoon Kim and Alexander~M. Rush. 2016.
\newblock \href {https://doi.org/10.18653/v1/d16-1139} {Sequence-level
  knowledge distillation}.
\newblock In \emph{Proceedings of the 2016 Conference on Empirical Methods in
  Natural Language Processing, {EMNLP} 2016, Austin, Texas, USA, November 1-4,
  2016}, pages 1317--1327. The Association for Computational Linguistics.

\bibitem[{Kirkpatrick et~al.(2017)Kirkpatrick, Pascanu, Rabinowitz, Veness,
  Desjardins, Rusu, Milan, Quan, Ramalho, Grabska-Barwinska, Hassabis, Clopath,
  Kumaran, and Hadsell}]{Kirkpatrick3521}
James Kirkpatrick, Razvan Pascanu, Neil Rabinowitz, Joel Veness, Guillaume
  Desjardins, Andrei~A. Rusu, Kieran Milan, John Quan, Tiago Ramalho, Agnieszka
  Grabska-Barwinska, Demis Hassabis, Claudia Clopath, Dharshan Kumaran, and
  Raia Hadsell. 2017.
\newblock \href {https://doi.org/10.1073/pnas.1611835114} {Overcoming
  catastrophic forgetting in neural networks}.
\newblock \emph{Proceedings of the National Academy of Sciences},
  114(13):3521--3526.

\bibitem[{Klein and Nabi(2021)}]{DBLP:journals/corr/abs-2109-05105}
Tassilo Klein and Moin Nabi. 2021.
\newblock Towards zero-shot commonsense reasoning with self-supervised
  refinement of language models.
\newblock \emph{CoRR}, abs/2109.05105.

\bibitem[{Lewis et~al.(2020)Lewis, Liu, Goyal, Ghazvininejad, Mohamed, Levy,
  Stoyanov, and Zettlemoyer}]{DBLP:conf/acl/LewisLGGMLSZ20}
Mike Lewis, Yinhan Liu, Naman Goyal, Marjan Ghazvininejad, Abdelrahman Mohamed,
  Omer Levy, Veselin Stoyanov, and Luke Zettlemoyer. 2020.
\newblock {BART:} denoising sequence-to-sequence pre-training for natural
  language generation, translation, and comprehension.
\newblock In \emph{{ACL}}, pages 7871--7880. Association for Computational
  Linguistics.

\bibitem[{Li et~al.(2019)Li, Chen, and Yu}]{DBLP:journals/corr/abs-1909-09743}
Shiyang Li, Jianshu Chen, and Dian Yu. 2019.
\newblock Teaching pretrained models with commonsense reasoning: {A}
  preliminary kb-based approach.
\newblock \emph{CoRR}, abs/1909.09743.

\bibitem[{Lin et~al.(2019)Lin, Chen, Chen, and Ren}]{DBLP:conf/emnlp/LinCCR19}
Bill~Yuchen Lin, Xinyue Chen, Jamin Chen, and Xiang Ren. 2019.
\newblock Kagnet: Knowledge-aware graph networks for commonsense reasoning.
\newblock In \emph{{EMNLP/IJCNLP} {(1)}}, pages 2829--2839. Association for
  Computational Linguistics.

\bibitem[{Lin et~al.(2020)Lin, Zhou, Shen, Zhou, Bhagavatula, Choi, and
  Ren}]{DBLP:conf/emnlp/LinZSZBCR20}
Bill~Yuchen Lin, Wangchunshu Zhou, Ming Shen, Pei Zhou, Chandra Bhagavatula,
  Yejin Choi, and Xiang Ren. 2020.
\newblock \href {https://doi.org/10.18653/v1/2020.findings-emnlp.165}
  {Commongen: {A} constrained text generation challenge for generative
  commonsense reasoning}.
\newblock In \emph{Findings of the Association for Computational Linguistics:
  {EMNLP} 2020, Online Event, 16-20 November 2020}, volume {EMNLP} 2020 of
  \emph{Findings of {ACL}}, pages 1823--1840. Association for Computational
  Linguistics.

\bibitem[{Liu et~al.(2020)Liu, Wan, He, Peng, and Yu}]{liu2020kgbart}
Ye~Liu, Yao Wan, Lifang He, Hao Peng, and Philip~S. Yu. 2020.
\newblock \href {http://arxiv.org/abs/2009.12677} {Kg-bart: Knowledge
  graph-augmented bart for generative commonsense reasoning}.

\bibitem[{Liu et~al.(2021)Liu, Wan, He, Peng, and
  Yu}]{DBLP:conf/aaai/LiuW0PY21}
Ye~Liu, Yao Wan, Lifang He, Hao Peng, and Philip~S. Yu. 2021.
\newblock {KG-BART:} knowledge graph-augmented {BART} for generative
  commonsense reasoning.
\newblock In \emph{{AAAI}}, pages 6418--6425. {AAAI} Press.

\bibitem[{Manning et~al.(2020)Manning, Clark, Hewitt, Khandelwal, and
  Levy}]{DBLP:journals/pnas/ManningCHKL20}
Christopher~D. Manning, Kevin Clark, John Hewitt, Urvashi Khandelwal, and Omer
  Levy. 2020.
\newblock \href {https://doi.org/10.1073/pnas.1907367117} {Emergent linguistic
  structure in artificial neural networks trained by self-supervision}.
\newblock \emph{Proc. Natl. Acad. Sci. {USA}}, 117(48):30046--30054.

\bibitem[{Merrill et~al.(2021)Merrill, Goldberg, Schwartz, and
  Smith}]{DBLP:journals/corr/abs-2104-10809}
William Merrill, Yoav Goldberg, Roy Schwartz, and Noah~A. Smith. 2021.
\newblock Provable limitations of acquiring meaning from ungrounded form: What
  will future language models understand?
\newblock \emph{CoRR}, abs/2104.10809.

\bibitem[{Mihaylov et~al.(2018)Mihaylov, Clark, Khot, and
  Sabharwal}]{mihaylov2018can}
Todor Mihaylov, Peter Clark, Tushar Khot, and Ashish Sabharwal. 2018.
\newblock Can a suit of armor conduct electricity? a new dataset for open book
  question answering.
\newblock \emph{arXiv preprint arXiv:1809.02789}.

\bibitem[{Peters et~al.(2019)Peters, Neumann, Logan~IV, Schwartz, Joshi, Singh,
  and Smith}]{peters2019knowledge}
Matthew~E Peters, Mark Neumann, Robert~L Logan~IV, Roy Schwartz, Vidur Joshi,
  Sameer Singh, and Noah~A Smith. 2019.
\newblock Knowledge enhanced contextual word representations.
\newblock \emph{arXiv preprint arXiv:1909.04164}.

\bibitem[{Petroni et~al.(2019)Petroni, Rockt{\"{a}}schel, Riedel, Lewis,
  Bakhtin, Wu, and Miller}]{DBLP:conf/emnlp/PetroniRRLBWM19}
Fabio Petroni, Tim Rockt{\"{a}}schel, Sebastian Riedel, Patrick S.~H. Lewis,
  Anton Bakhtin, Yuxiang Wu, and Alexander~H. Miller. 2019.
\newblock \href {https://doi.org/10.18653/v1/D19-1250} {Language models as
  knowledge bases?}
\newblock In \emph{Proceedings of the 2019 Conference on Empirical Methods in
  Natural Language Processing and the 9th International Joint Conference on
  Natural Language Processing, {EMNLP-IJCNLP} 2019, Hong Kong, China, November
  3-7, 2019}, pages 2463--2473. Association for Computational Linguistics.

\bibitem[{Raffel et~al.(2019)Raffel, Shazeer, Roberts, Lee, Narang, Matena,
  Zhou, Li, and Liu}]{raffel2019exploring}
Colin Raffel, Noam Shazeer, Adam Roberts, Katherine Lee, Sharan Narang, Michael
  Matena, Yanqi Zhou, Wei Li, and Peter~J Liu. 2019.
\newblock Exploring the limits of transfer learning with a unified text-to-text
  transformer.
\newblock \emph{arXiv preprint arXiv:1910.10683}.

\bibitem[{Raffel et~al.(2020)Raffel, Shazeer, Roberts, Lee, Narang, Matena,
  Zhou, Li, and Liu}]{DBLP:journals/jmlr/RaffelSRLNMZLL20}
Colin Raffel, Noam Shazeer, Adam Roberts, Katherine Lee, Sharan Narang, Michael
  Matena, Yanqi Zhou, Wei Li, and Peter~J. Liu. 2020.
\newblock \href {http://jmlr.org/papers/v21/20-074.html} {Exploring the limits
  of transfer learning with a unified text-to-text transformer}.
\newblock \emph{J. Mach. Learn. Res.}, 21:140:1--140:67.

\bibitem[{Rajpurkar et~al.(2016)Rajpurkar, Zhang, Lopyrev, and Liang}]{qnli}
Pranav Rajpurkar, Jian Zhang, Konstantin Lopyrev, and Percy Liang. 2016.
\newblock Squad: 100, 000+ questions for machine comprehension of text.
\newblock In \emph{{EMNLP}}.

\bibitem[{Roberts et~al.(2020)Roberts, Raffel, and
  Shazeer}]{DBLP:conf/emnlp/RobertsRS20}
Adam Roberts, Colin Raffel, and Noam Shazeer. 2020.
\newblock \href {https://doi.org/10.18653/v1/2020.emnlp-main.437} {How much
  knowledge can you pack into the parameters of a language model?}
\newblock In \emph{Proceedings of the 2020 Conference on Empirical Methods in
  Natural Language Processing, {EMNLP} 2020, Online, November 16-20, 2020},
  pages 5418--5426. Association for Computational Linguistics.

\bibitem[{Roemmele et~al.(2011)Roemmele, Bejan, and
  Gordon}]{DBLP:conf/aaaiss/RoemmeleBG11}
Melissa Roemmele, Cosmin~Adrian Bejan, and Andrew~S. Gordon. 2011.
\newblock \href {http://www.aaai.org/ocs/index.php/SSS/SSS11/paper/view/2418}
  {Choice of plausible alternatives: An evaluation of commonsense causal
  reasoning}.
\newblock In \emph{Logical Formalizations of Commonsense Reasoning, Papers from
  the 2011 {AAAI} Spring Symposium, Technical Report SS-11-06, Stanford,
  California, USA, March 21-23, 2011}. {AAAI}.

\bibitem[{Sap et~al.(2019{\natexlab{a}})Sap, Bras, Allaway, Bhagavatula,
  Lourie, Rashkin, Roof, Smith, and Choi}]{DBLP:conf/aaai/SapBABLRRSC19}
Maarten Sap, Ronan~Le Bras, Emily Allaway, Chandra Bhagavatula, Nicholas
  Lourie, Hannah Rashkin, Brendan Roof, Noah~A. Smith, and Yejin Choi.
  2019{\natexlab{a}}.
\newblock {ATOMIC:} an atlas of machine commonsense for if-then reasoning.
\newblock In \emph{{AAAI}}, pages 3027--3035. {AAAI} Press.

\bibitem[{Sap et~al.(2019{\natexlab{b}})Sap, Rashkin, Chen, {Le Bras}, and
  Choi}]{sap2019socialiqa}
Maarten Sap, Hannah Rashkin, Derek Chen, Ronan {Le Bras}, and Yejin Choi.
  2019{\natexlab{b}}.
\newblock Socialiqa: Commonsense reasoning about social interactions.
\newblock In \emph{{EMNLP}}.

\bibitem[{Socher et~al.(2013)Socher, Perelygin, Wu, Chuang, Manning, Ng, and
  Potts}]{sst}
Richard Socher, Alex Perelygin, Jean Wu, Jason Chuang, Christopher~D. Manning,
  Andrew~Y. Ng, and Christopher Potts. 2013.
\newblock Recursive deep models for semantic compositionality over a sentiment
  treebank.
\newblock In \emph{{EMNLP}}.

\bibitem[{Speer et~al.(2017)Speer, Chin, and Havasi}]{DBLP:conf/aaai/SpeerCH17}
Robyn Speer, Joshua Chin, and Catherine Havasi. 2017.
\newblock Conceptnet 5.5: An open multilingual graph of general knowledge.
\newblock In \emph{{AAAI}}, pages 4444--4451. {AAAI} Press.

\bibitem[{Talmor et~al.(2018)Talmor, Herzig, Lourie, and
  Berant}]{talmor2018commonsenseqa}
Alon Talmor, Jonathan Herzig, Nicholas Lourie, and Jonathan Berant. 2018.
\newblock Commonsenseqa: A question answering challenge targeting commonsense
  knowledge.
\newblock \emph{arXiv preprint arXiv:1811.00937}.

\bibitem[{Tu et~al.(2020)Tu, Lalwani, Gella, and
  He}]{DBLP:journals/tacl/TuLGH20}
Lifu Tu, Garima Lalwani, Spandana Gella, and He~He. 2020.
\newblock \href {https://transacl.org/ojs/index.php/tacl/article/view/2055} {An
  empirical study on robustness to spurious correlations using pre-trained
  language models}.
\newblock \emph{Trans. Assoc. Comput. Linguistics}, 8:621--633.

\bibitem[{Vaswani et~al.(2017)Vaswani, Shazeer, Parmar, Uszkoreit, Jones,
  Gomez, Kaiser, and Polosukhin}]{DBLP:conf/nips/VaswaniSPUJGKP17}
Ashish Vaswani, Noam Shazeer, Niki Parmar, Jakob Uszkoreit, Llion Jones,
  Aidan~N. Gomez, Lukasz Kaiser, and Illia Polosukhin. 2017.
\newblock Attention is all you need.
\newblock In \emph{{NIPS}}, pages 5998--6008.

\bibitem[{Wang et~al.(2019)Wang, Singh, Michael, Hill, Levy, and Bowman}]{glue}
Alex Wang, Amanpreet Singh, Julian Michael, Felix Hill, Omer Levy, and
  Samuel~R. Bowman. 2019.
\newblock {GLUE:} {A} multi-task benchmark and analysis platform for natural
  language understanding.
\newblock In \emph{{ICLR}}.

\bibitem[{Wang et~al.(2020)Wang, Tang, Duan, Wei, Huang, Cao, Jiang, Zhou
  et~al.}]{wang2020k}
Ruize Wang, Duyu Tang, Nan Duan, Zhongyu Wei, Xuanjing Huang, Cuihong Cao,
  Daxin Jiang, Ming Zhou, et~al. 2020.
\newblock K-adapter: Infusing knowledge into pre-trained models with adapters.
\newblock \emph{arXiv preprint arXiv:2002.01808}.

\bibitem[{Warstadt et~al.(2019)Warstadt, Singh, and Bowman}]{cola}
Alex Warstadt, Amanpreet Singh, and Samuel~R. Bowman. 2019.
\newblock Neural network acceptability judgments.
\newblock \emph{{TACL}}.

\bibitem[{Williams et~al.(2018)Williams, Nangia, and Bowman}]{mnli}
Adina Williams, Nikita Nangia, and Samuel~R. Bowman. 2018.
\newblock A broad-coverage challenge corpus for sentence understanding through
  inference.
\newblock In \emph{{NAACL-HLT}}.

\bibitem[{Wolf et~al.(2019)Wolf, Debut, Sanh, Chaumond, Delangue, Moi, Cistac,
  Rault, Louf, Funtowicz, Davison, Shleifer, von Platen, Ma, Jernite, Plu, Xu,
  Scao, Gugger, Drame, Lhoest, and Rush}]{Wolf2019HuggingFacesTS}
Thomas Wolf, Lysandre Debut, Victor Sanh, Julien Chaumond, Clement Delangue,
  Anthony Moi, Pierric Cistac, Tim Rault, Rémi Louf, Morgan Funtowicz, Joe
  Davison, Sam Shleifer, Patrick von Platen, Clara Ma, Yacine Jernite, Julien
  Plu, Canwen Xu, Teven~Le Scao, Sylvain Gugger, Mariama Drame, Quentin Lhoest,
  and Alexander~M. Rush. 2019.
\newblock Huggingface's transformers: State-of-the-art natural language
  processing.
\newblock \emph{ArXiv}, abs/1910.03771.

\bibitem[{Xiong et~al.(2020)Xiong, Du, Wang, and
  Stoyanov}]{Xiong2020Pretrained}
Wenhan Xiong, Jingfei Du, William~Yang Wang, and Veselin Stoyanov. 2020.
\newblock \href {https://openreview.net/forum?id=BJlzm64tDH} {Pretrained
  encyclopedia: Weakly supervised knowledge-pretrained language model}.
\newblock In \emph{International Conference on Learning Representations}.

\bibitem[{Xu et~al.(2021)Xu, Zhou, Ge, Xu, McAuley, and
  Wei}]{DBLP:conf/naacl/XuZGXMW21}
Canwen Xu, Wangchunshu Zhou, Tao Ge, Ke~Xu, Julian~J. McAuley, and Furu Wei.
  2021.
\newblock Blow the dog whistle: {A} chinese dataset for cant understanding with
  common sense and world knowledge.
\newblock In \emph{{NAACL-HLT}}, pages 2139--2145. Association for
  Computational Linguistics.

\bibitem[{Ye et~al.(2019)Ye, Chen, Wang, and
  Ling}]{DBLP:journals/corr/abs-1908-06725}
Zhi{-}Xiu Ye, Qian Chen, Wen Wang, and Zhen{-}Hua Ling. 2019.
\newblock \href {http://arxiv.org/abs/1908.06725} {Align, mask and select: {A}
  simple method for incorporating commonsense knowledge into language
  representation models}.
\newblock \emph{CoRR}, abs/1908.06725.

\bibitem[{Zhang et~al.(2019)Zhang, Han, Liu, Jiang, Sun, and
  Liu}]{zhang2019ernie}
Zhengyan Zhang, Xu~Han, Zhiyuan Liu, Xin Jiang, Maosong Sun, and Qun Liu. 2019.
\newblock Ernie: Enhanced language representation with informative entities.
\newblock \emph{arXiv preprint arXiv:1905.07129}.

\bibitem[{Zhou et~al.(2021)Zhou, Lee, Selvam, Lee, and
  Ren}]{DBLP:conf/iclr/ZhouLSL021}
Wangchunshu Zhou, Dong{-}Ho Lee, Ravi~Kiran Selvam, Seyeon Lee, and Xiang Ren.
  2021.
\newblock Pre-training text-to-text transformers for concept-centric common
  sense.
\newblock In \emph{{ICLR}}. OpenReview.net.

\bibitem[{Zhou et~al.(2020{\natexlab{a}})Zhou, Zhang, Cui, and
  Huang}]{DBLP:conf/aaai/ZhouZCH20}
Xuhui Zhou, Yue Zhang, Leyang Cui, and Dandan Huang. 2020{\natexlab{a}}.
\newblock Evaluating commonsense in pre-trained language models.
\newblock In \emph{{AAAI}}, pages 9733--9740. {AAAI} Press.

\bibitem[{Zhou et~al.(2020{\natexlab{b}})Zhou, Zhang, Cui, and
  Huang}]{zhou2020evaluating}
Xuhui Zhou, Yue Zhang, Leyang Cui, and Dandan Huang. 2020{\natexlab{b}}.
\newblock Evaluating commonsense in pre-trained language models.
\newblock In \emph{AAAI}, pages 9733--9740.

\bibitem[{Zhu et~al.(2015)Zhu, Kiros, Zemel, Salakhutdinov, Urtasun, Torralba,
  and Fidler}]{DBLP:conf/iccv/ZhuKZSUTF15}
Yukun Zhu, Ryan Kiros, Richard~S. Zemel, Ruslan Salakhutdinov, Raquel Urtasun,
  Antonio Torralba, and Sanja Fidler. 2015.
\newblock \href {https://doi.org/10.1109/ICCV.2015.11} {Aligning books and
  movies: Towards story-like visual explanations by watching movies and reading
  books}.
\newblock In \emph{2015 {IEEE} International Conference on Computer Vision,
  {ICCV} 2015, Santiago, Chile, December 7-13, 2015}, pages 19--27. {IEEE}
  Computer Society.

\end{thebibliography}

\newpage

\begin{table*}[!h]
\centering
	\centering
	\resizebox{\textwidth}{!}{
		\begin{tabular}{lccccccccc}
			\toprule
			\textbf{Methods} &
            \textbf{CoLA} &  \textbf{MNLI} & \textbf{MRPC} & \textbf{QNLI}  & \textbf{QQP}   & \textbf{RTE} &  \textbf{SST-2} & \textbf{SST-B}  & \textbf{Meta Score} \\
			\midrule
			BERT-base & 58.9 & 84.7 & 89.6 & 91.2 & 90.0 & 71.4 & 93.0 & 90.0 & 83.6  \\ 
			T5-base & 55.9 & 84.5 & 90.3 & 90.5 & 90.2 & 76.2 & 92.8 & 87.8 & 83.5 \\
            \baby-base & 57.4 & 84.4 & 90.6 & 90.9 & 89.9 & 76.8 & 92.5 & 88.4 & 83.9 \\ 
            
			\bottomrule
	\end{tabular}
	}
	\caption{Experimental results of base-size models on the GLUE benchmark.} 
	\vspace{-0.4cm}
	\label{tab:result-glue}
\end{table*}

\appendix

\section{Pre-training and Fine-tuning Details}

\subsection{Pre-Training Details}
\label{appendix:pretrain}
We implement our models using Pytorch-lightning~\citep{falcon2019pytorch} and Hugginface's Pytorch Transformers~\citep{Wolf2019HuggingFacesTS}.
For pre-training phase, we use the AdamW optimizer with maximum sequence length 256, train batch size 8, gradient accumulation 8, warmup steps 8000, weight decay 0.01 and adam epsilon 1e-6. We train the models with 8 V100 GPUs and FP32 precision. The model is pre-trained for 10 epochs.
We searched for the best learning rate for our model out of [5e-6, 2e-5, 5e-5, 1e-4].

\subsection{Fine-Tuning Details}
\label{appendix:finetune}
For fine-tuning, we use 4 V100 GPUs and use FP32.
For all tasks, we use the AdamW optimizer with learning rate from [1e-5, 2e-5, 5e-5, 1e-4, 2e-4], maximum sequence length 256, batch size from [4, 8, 16, 32]. For all tasks, we use a warmup fraction of 0.01, and max epoch of 20.

\section{Additional Analysis}

\subsection{Qualitative Analysis} 
\begin{figure}
    \centering
    \includegraphics[width=\linewidth]{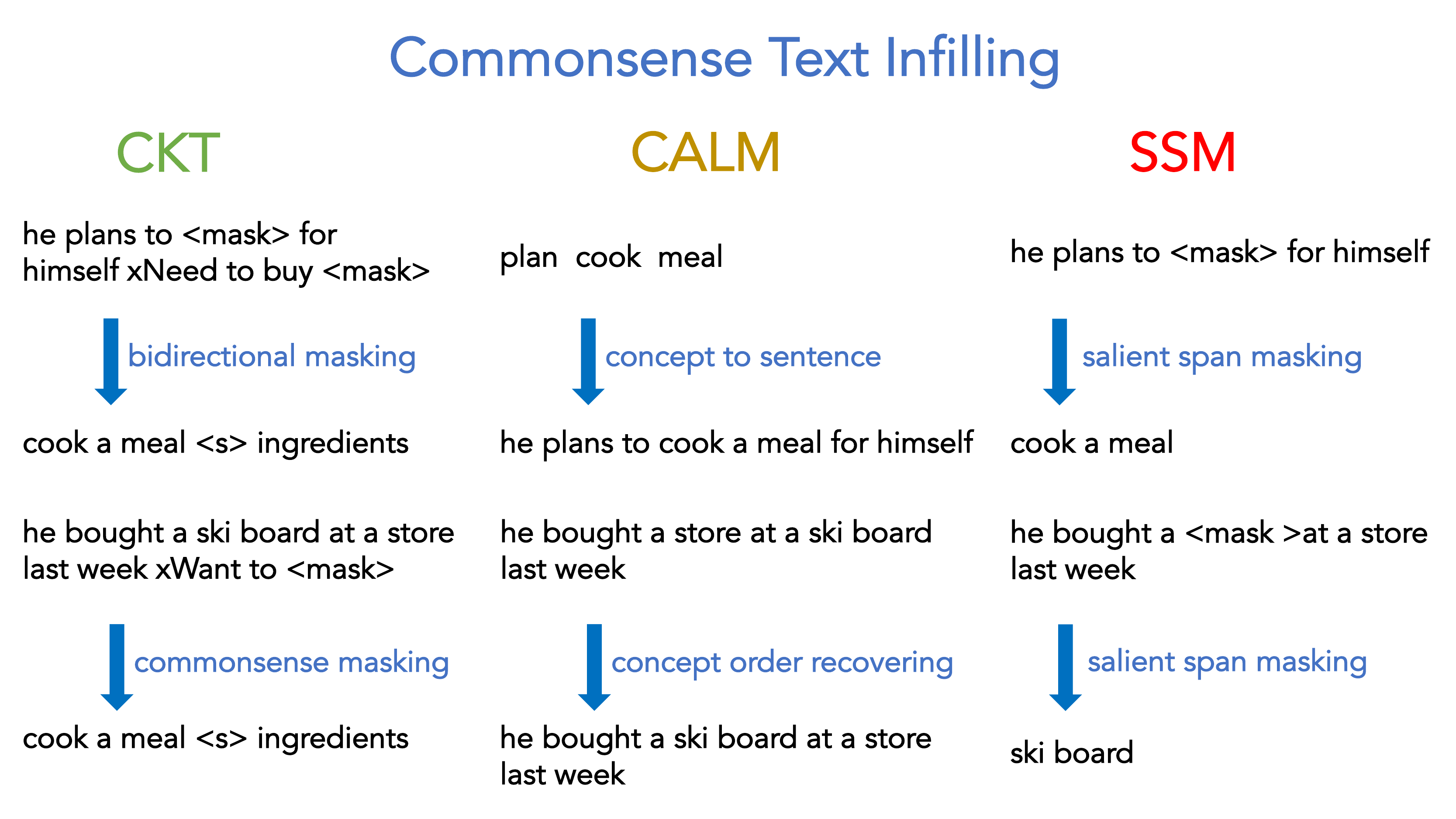}
    \caption{Performance on the CSQA dataset w.r.t. the size of training data used for \method.}
    \label{fig:casestudy}
\end{figure}

To better understand the proposed method, we present a case study in Figure \ref{fig:casestudy}. We can see that both the objectives introduced in the CALM model and the salient span masking (SSM) strategy fail to exploit the underlying commonsense rationale beyond the surface form of texts while our approach directly aligns texts with the corresponding commonsense inferences with different commonsense relations. That explains why \method can effectively improve a pre-trained model's performance on downstream tasks requiring commonsense reasoning ability.

\subsection{Experimental Results on GLUE} To verify that \method is suitable for general-purpose pre-trained models, we fine-tune our model on the GLUE benchmark~\citep{glue}. Specifically, we test on MRPC~\citep{mrpc}, QQP\footnote{\url{https://www.quora.com/q/quoradata/First-Quora-Dataset-Release-Question-Pairs}} and STS-B~\citep{senteval} for Paraphrase Similarity Matching; SST-2~\citep{sst} for Sentiment Classification; MNLI~\citep{mnli}, QNLI~\citep{qnli} and RTE~\citep{glue} for the Natural Language Inference; CoLA~\citep{cola} for Linguistic Acceptability.

The results are shown on Table \ref{tab:result-glue}, we can see that after \method, the resulting model's general natural language understanding ability is comparable with the original T5-base model. This shows that our approach does not affect the model's general transfer ability and thus can be applied to general-purpose language models.

\begin{figure}
    \centering
    \includegraphics[width=\linewidth]{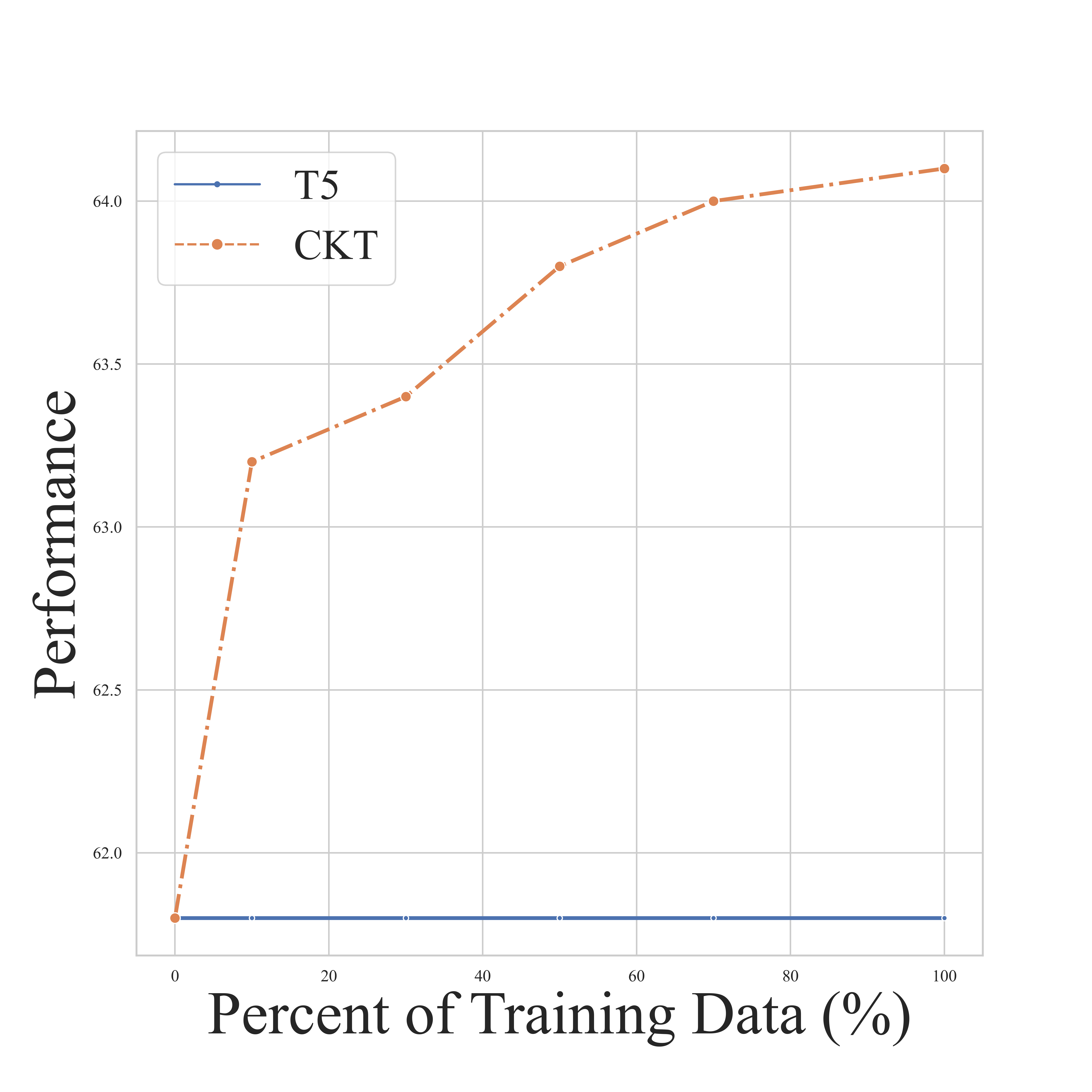}
    \caption{Performance on the CSQA dataset w.r.t. the size of training data used for \method.}
    \label{fig:datasize}
\end{figure}

\begin{table*}[!h]
\centering
	\centering
	\resizebox{0.9\textwidth}{!}{
		\begin{tabular}{lcccccc}
			\toprule
			\textbf{Methods} &
            \textbf{CSQA} &  \textbf{OBQA} & \textbf{PIQA} & \textbf{aNLI} & \textbf{\textsc{SocialIQA}} & \textbf{COPA} \\
			\midrule
			BART & 72.31 & 65.80 & 74.12 & 78.27 & 71.6 & 85.6 \\
            \baby-BART & 73.14 & 68.20 & 76.95 & 79.52 & 73.3 & 87.2\\ 
			\bottomrule
	\end{tabular}
	}
	\caption{Experimental results (mean of 3 random runs) with BART.} 
	\vspace{-0.5cm}
	\label{tab:result-bart}
\end{table*}

\subsection{Experiments with BART} To demonstrate the versatility of \method for different backbones, we conduct additional experiments using BART as the backbone model. The results are shown in Table \ref{tab:result-bart}. We can see that \method also consistently improves the BART model, demonstrating the versatility of our approach.

\begin{table}[!h]
\centering
\resizebox{\linewidth}{!}{
		\begin{tabular}{lcccc}
			\toprule
			\textbf{Methods} &
            \textbf{BLEU-4} &  \textbf{METEOR} & \textbf{CIDEr} & \textbf{SPICE} \\
			\midrule
			\bf T5-base & 24.90 & 31.20 & 12.99 & 32.40 \\
			\bf CALM-base & \bf 26.40 & \bf 31.40 & \bf 13.88 & 33.00 \\
			\bf CKT-base & 26.20 & \bf 31.40 & 13.65 & \bf 33.10 \\
			\bottomrule
	\end{tabular}
	}
	\caption{Experimental results (mean of 3 random runs) with BART.} 
	\vspace{-0.2cm}
	\label{tab:result-commongen}
\end{table}

\subsection{Experiments on CommonGEN} We also experiment on the CommonGEN dataset, a generative commonsense reasoning dataset where the model is required to take several keywords as inputs and output a sentence that makes sense. The results are shown in Table \ref{tab:result-commongen}. We can see that our approach performs similarly with the CALM model, which includes the CommonGEN task objective as one of the pre-training tasks.

\subsection{Impact of Pre-training Data Size} We also conduct experiments to investigate the sample-efficiency of \method. We present the trend of performance improvement in Figure \ref{fig:datasize}. We can see that our method achieves significant performance improvement upon the T5 baseline with only 10\% of the total training data, which confirms the sample-efficiency of \method.


\end{document}